\documentclass[review]{elsarticle}
\usepackage{xpatch}
\usepackage{array}
\newcolumntype{P}[1]{>{\centering\arraybackslash}m{#1}}
\usepackage[table]{xcolor} % Needed for cell coloring

\usepackage{booktabs} % for table rules
\usepackage{array} % for new column types

\usepackage{multirow}
\usepackage{amsmath,amssymb,amsfonts}
\usepackage{algorithm}
\usepackage{algpseudocode}
\usepackage{graphicx}
\usepackage{textcomp}
\usepackage{xcolor}
\usepackage[flushleft]{threeparttable}
\usepackage{enumitem}
\usepackage{{booktabs}}
\usepackage[bottom]{footmisc}
\usepackage{longtable}
\usepackage{caption}
\usepackage{subcaption} % for subtables
\usepackage{afterpage}

\usepackage{url}
\usepackage[hidelinks]{hyperref}
\xpatchcmd{\State}{\algorithmicend\ \algorithmicfor}{\algorithmicend}{}{}

\usepackage{lineno,hyperref}
\usepackage{float} % For the [H] placement option
\usepackage{autonum}
\usepackage{cleveref}
\makeatletter
\def\ps@pprintTitle{%
 \let\@oddhead\@empty
 \let\@evenhead\@empty
 \def\@oddfoot{\reset@font\hfil\thepage\hfil}
 \let\@evenfoot\@oddfoot
}
\makeatother
\begin{document}

\begin{frontmatter}

\title{Federated Transfer Learning with Task Personalization for Condition Monitoring in Ultrasonic Metal Welding}
% \tnotetext[mytitlenote]{Fully documented templates are available in the elsarticle package on \href{http://www.ctan.org/tex-archive/macros/latex/contrib/elsarticle}{CTAN}.}

% %% Group authors per affiliation:
% \address{Radarweg 29, Amsterdam}
% \fntext[myfootnote]{Since 1880.}

% %% or include affiliations in footnotes:
% \author[mymainaddress,mysecondaryaddress]{Elsevier Inc}

% \author[mysecondaryaddress]{Global Customer Service\corref{mycorrespondingauthor}}
% \cortext[mycorrespondingauthor]{Corresponding author}

% \address[mymainaddress]{1600 John F Kennedy Boulevard, Philadelphia}
% \address[mysecondaryaddress]{360 Park Avenue South, New York}
\author[UIUCMechSE,UIUCCSL]{Ahmadreza Eslaminia}
\ead{ae15@illinois.edu}

\author[UIUCMechSE]{Yuquan Meng}
\ead{yuquanm2@illinois.edu}

\author[UIUCCSL]{Klara Nahrstedt}
\ead{klara@illinois.edu}

\author[UIUCMechSE,UMich]{Chenhui Shao\corref{mycorrespondingauthor}}
\ead{chshao@umich.edu}

\address[UIUCMechSE]{Department of Mechanical Science and Engineering, University of Illinois at Urbana-Champaign, Urbana, IL 61801, USA}
\address[UIUCCSL]{Coordinated Science Laboratory, University of Illinois at Urbana-Champaign, Urbana, IL 61801, USA}
\address[UMich]{Department of Mechanical Engineering, University of Michigan, Ann Arbor, MI 48109, USA}

\cortext[mycorrespondingauthor]{Corresponding author}

\begin{abstract}
Ultrasonic metal welding (UMW) is a key joining technology with widespread industrial applications. Condition monitoring (CM) capabilities are critically needed in UMW applications, because process anomalies, such as tool degradation and workpiece surface contamination, significantly deteriorate the joining quality. Recently, machine learning models emerged as a promising tool for CM in many manufacturing applications due to their ability to learn complex patterns. Yet, the successful deployment of these models requires substantial training data that may be expensive and time-consuming to collect. Additionally, many existing machine learning models lack generalizability or adaptability and cannot be directly applied to new process configurations (i.e., domains). Such issues may be potentially alleviated by pooling data across manufacturers or factories, but data sharing raises critical data privacy concerns that have prohibited collaborative learning in industry. To address these challenges, this paper presents a Federated Transfer Learning with Task Personalization (FTL-TP) framework that provides domain generalization capabilities in distributed learning while ensuring data privacy. By effectively learning a unified representation from feature space, FTL-TP can adapt CM models for clients working on similar tasks, thereby enhancing their overall adaptability and performance jointly. To demonstrate the effectiveness of FTL-TP, we investigate two distinct UMW CM tasks, including tool condition monitoring and workpiece surface condition classification. Compared with state-of-the-art FL algorithms, FTL-TP achieves a 5.35\%--8.08\% improvement of accuracy in CM in new target domains. FTL-TP is also shown to achieve excellent performance in challenging scenarios involving unbalanced data distributions and limited client fractions. Furthermore, by implementing the FTL-TP method on an edge-cloud architecture, we show that this method is both viable and efficient in practice. The FTL-TP framework is readily extensible to various other manufacturing applications.

\end{abstract}


\end{frontmatter}

% \linenumbers

\section{Introduction}\label{sec:introduction}

Ultrasonic metal welding (UMW) is an advanced joining technology, which utilizes high-frequency oscillation shear forces to create strong solid-state joint. It has been used in a wide range of industrial applications, including automotive body construction~\cite{ni2018ultrasonic,de2022review} and lithium-ion battery assembly~\cite{caiUSW,balz2020process}. Compared to conventional fusion welding methods, UMW offers several important advantages, such as reduced energy consumption, lower emissions, higher production rates, and environmental friendliness~\cite{meng2022physics}. Therefore, it is one of the promising dissimilar metal joining techniques for sustainable manufacturing.

Despite numerous advantages, the quality of UMW joint is significantly influenced by process anomalies that may take different forms. Tool degradation and surface contamination of workpieces are two major causes of such anomalies in UMW~\cite{shao2016tool,nunes2022influence,nazir2021online}. As such, developing effective and efficient condition monitoring (CM) method is of vital importance to ensure quality and reliability of industrial UMW. CM for UMW has attracted in-depth research attention~\cite{shao2016tool,nazir2021online}. Some methods achieved close-to-perfect CM performance when sufficient training data is available~\cite{nazir2021online,lu2023online}.

Nevertheless, the existing CM methods are mostly created for a fixed configuration (i.e., domain). When the UMW configurations, such as material of specimens and welding parameters, are changed, the well-trained models do not work well and their accuracy substantially drops~\cite{meng2023explainable}. To adapt to new process configurations, existing CM methods require access to sufficient training data collected from the new configurations. However, re-collection of such data can be both expensive and time-consuming in manufacturing \cite{mehta2021adaptive}. This necessitates the development of a methodology that can generalize and adapt to new configurations efficiently.

Previous studies showed that having access to extensive data encompassing diverse configurations/domains may improve a model's generalizability, which implies that pooling data from diverse configurations provides a potential solution~\cite{mehta2022federated}. However, a single company or factory may not possess enough data covering multiple configurations. Consequently, collaborative model training across different companies/factories could be advantageous. However, collaborative learning through data pooling may raise significant data privacy concerns \cite{mehta2023federated}. Furthermore, resistance to data integration can occur even within different departments of the same organization~\cite{yang2019federated}. 

This paper aims to develop a method with high generalizability to unseen domains through collaborative data analysis across different manufacturing sites without compromising data privacy. To this end, an innovative Federated Transfer Learning with Task Personalization (FTL-TP) framework is proposed. FTL-TP is designed to operate across two or more distinct UMW CM tasks, such as tool condition monitoring and surface condition classification. These tasks are explored under various process configurations/domains, such as different materials for the joining sheets. This study introduces a strategy to effectively transfer knowledge by learning a generalized representation of the feature space shared across different datasets acquired from different companies while still preserving privacy. Additionally, we introduce a new loss function to improve both the generalizability and convergence of the model under varying configurations/domains. The main contributions of this research are summarized as follows:
\begin{enumerate}
\item The FTL-TP framework successfully achieves domain generalization in the context of FL. Compared with baseline FL algorithms, such as FedAvg and FedProx, our framework exhibits superior performance, with an accuracy improvement of 5.35\%--8.08\% in new, previously unseen target domains. 

\item By implementing FL, this research is the first to integrate data privacy into data-driven CM of UMW. This addresses the industrial-wide concerns of data safety and data privacy, laying a foundation of large-scale collaborations among companies that use UMW in their production.

\item  To demonstrate the industrial applicability of the FTL-TP, we implement FTL-TP in a real-world cyber-manufacturing environment and evaluate the computing efficiency. By running on an edge-cloud architecture, where Raspberry Pi’s serve as edge devices, and a message broker is utilized for weight transmission, FTL-TP is shown to be not only data-efficient but also time-efficient, proving its potential for widespread applications in smart manufacturing.

\end{enumerate}

The reminder of this paper is organized as follows. In Section~\ref{sec:related}, the existing literature is reviewed in detail. Section~\ref{sec:problem} formulates the problem and describes the dataset used in this study. Section~\ref{sec:method} introduces the proposed FTL-TP framework and presents its architecture and the algorithm in detail. Several case studies are presented in Section~\ref{sec:case} to thoroughly evaluate the effectiveness of the FTL-TP framework. Subsequently, Section~\ref{sec:discussion} provides an in-depth discussion about the learning process, implications of the findings, and directions for future research. Finally, Section~\ref{sec:conclusion} concludes the paper.

\section{Related Work and Background}\label{sec:related}

\subsection{Data-driven CM for UMW}

Data-driven CM methods are becoming increasingly crucial in many industrial applications \cite{wang2006condition,zhong2017iot,zhao2019deep,huang2024cross}, such as additive manufacturing, machining, and air brake systems ~\cite{dominguez2022machine, ferreira2022remaining, hou2023fault,yin2023online}. Similarly, in recent years, machine learning methods for monitoring UMW processes have also yielded promising outcomes \cite{nazir2021online,wu2022end,lu2023online}. For example, Nazir and Shao~\cite{nazir2021online} developed an online sensing system and investigated several machine learning algorithms, such as logistic regression, K-nearest neighbors, and SVM, to predict tool conditions in a real-time fashion. Wu et al.~\cite{wu2022end} utilized residual networks to predict joint quality in UMW with sensing signals. Schwarz et al.~\cite{schwarz2022improving} improved process monitoring of UMW by using linear regression as well as multi-layer perceptron (MLP) regression to predict tensile shear strength. Most recently, Lu et al.~\cite{lu2023online} implemented an MLP classifier for identifying mixed welding disturbances, focusing on tool conditions and material surface conditions.

Despite these advances, a common challenge among most of these studies is their limited generalization ability across varying industrial situations due to domain shift, which means the data distribution changes between different domains used for training and deployment. This discrepancy can significantly impact the performance of machine learning models under different configurations, as demonstrated in \cite{meng2023explainable,russell2023maximizing,huang2024cross}. Only a few studies have aimed at addressing this challenge for the CM of UMW processes. Tian et al.~\cite{tian2023weldmon} introduced a neural network (NN) model that incorporates an augmentation strategy to tackle the issue of domain shifts in CM tasks within UMW. Meng et al.~\cite{meng2023explainable} developed a few-shot learning method to address the variability of configurations in online anomaly detection for UMW. Despite good performance, these models encounter difficulties when there are no data available at all for target domains. This type of extreme data scarcity situations is referred to as ``domain generalization.''

Collaborative learning across data resources emerges as a viable solution to overcome data constraints, offering a potential remedy in such scenarios. Nonetheless, the implementation of collaborative learning often encounters privacy concerns, posing a significant barrier in many cases.

\subsection{Federated learning}
\begin{figure}[h]
\centering
\includegraphics[width=0.7\linewidth]{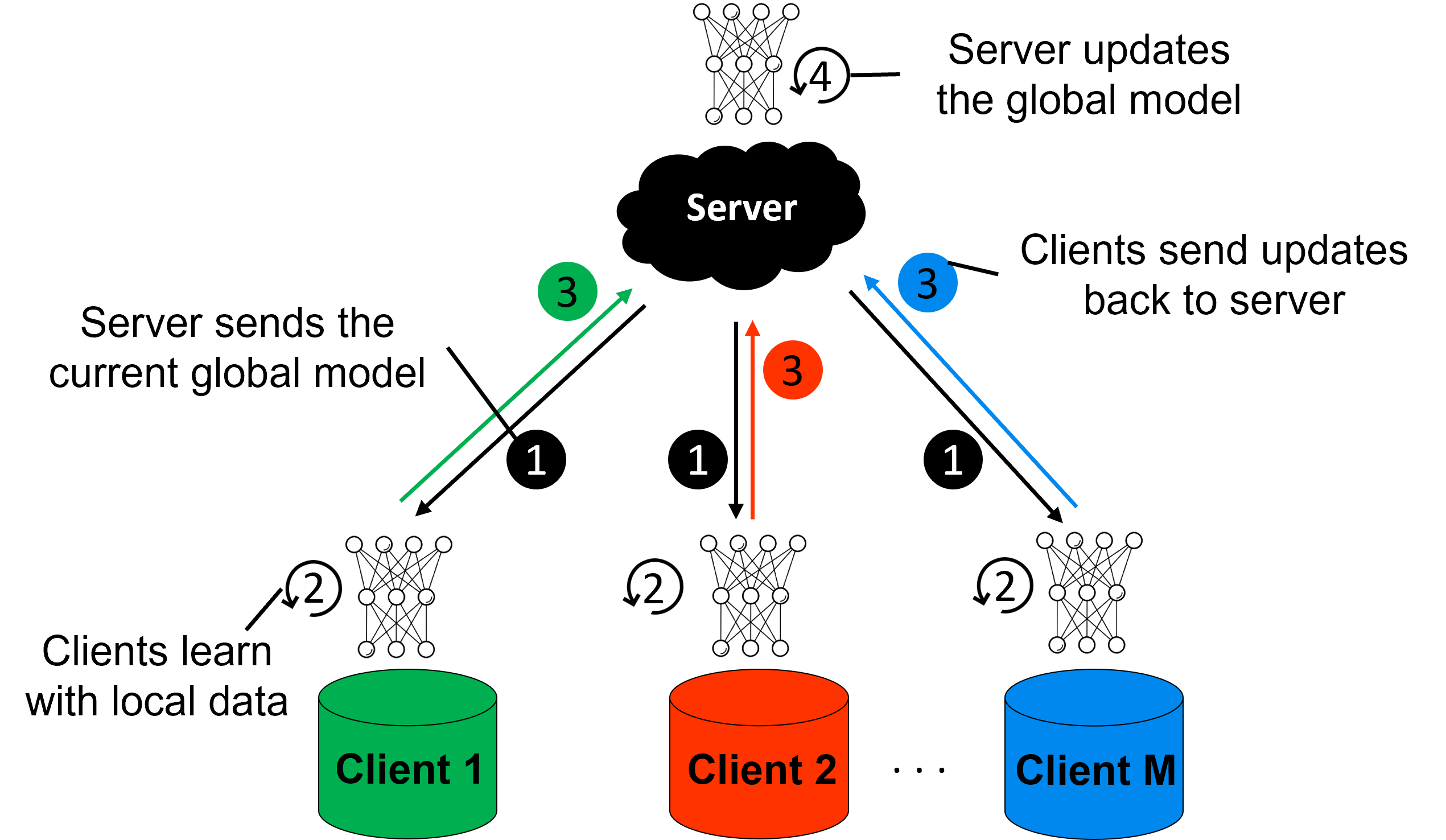}
\caption{A schematic for training process in FL regime.}
\label{fig:FL}
\end{figure}

In recent years, FL has emerged as a promising solution to the challenges of data availability and privacy in distributed learning environments \cite{hao2019efficient,li2020review,khan2020federated}. The FedAvg method, a widely recognized baseline introduced by McMahan et al.~\cite{mcmahan2017communication}, facilitates collaborative training across multiple data resources/clients without sharing raw data. Figure~\ref{fig:FL} illustrates a typical FL procedure. In each server round, the central server initially sends the model parameters to each participating client. Subsequently, clients independently train the received model on their local data and then transmit their local models' parameter updates back to the central server, which then aggregates these updates. The server round iterates until the model meets convergence criteria or a predetermined number of iterations is achieved.

The FedAvg algorithm assumes in each server round, we have \(m\) clients available from all \(M\) clients. The total number of data points \( N \) is obtained by summing \( n_k \), the data points from each of the \( m \) clients.
The primary objective of FedAvg is to minimize the following optimization function:

\begin{equation}
\arg\min_{\theta} F(\theta) = \frac{1}{N} \sum_{k=1}^m n_k \cdot F_k(\theta),
\end{equation}
where $\theta$ represents the model parameters,  and $F_k(\theta)$ corresponds to the loss function of the $k$-th client.

To effectively utilize FedAvg, it is essential to define some key parameters: client fraction \(C\), indicating the proportion of clients participating in each aggregation; local epoch \(E\), the number of training epochs per client per server round. Algorithm~\ref{alg:fedavg} in \ref{appendix:first} outlines the specific steps involved for both the client and server components of FedAvg.

To enhance the convergence of FedAvg's, Li et al. (2020)~\cite{li2020federated} proposed the FedProx method by adding a proximal term to the client loss, thereby limiting the impact of variable local updates. The objective function of each client in FedProx is:
\begin{equation}
\min_{\theta} h_k(\theta; \theta_t) = F_k(\theta) + \frac{\mu}{2} \|\theta - \theta_t\|^2,
\end{equation}
where $\mu$ represents a parameter controlling the proximal term, and $\|\theta - \theta_t\|^2$ denotes the squared norm of the difference between the current model $\theta$ and the previous model $\theta_t$. A large $\mu$ may lead to slower convergence by keeping updates near the initial point, while a small $\mu$ typically has little effect on convergence.

The model's ability to generalize to new users is paramount in FL. In this regard, Nguyen et al. (2022) ~\cite{nguyen2022fedsr} developed the FedL2R algorithm that incorporates an L2-norm regularize (L2R) into the representation of the network to improve the generalizability across new target domains by limiting the representation's information capacity. Thus, the final local objective function of each client \(k\) is:
\begin{equation}
\min_{\theta} F_k(\theta) + \alpha^{L2R} \cdot \mathcal{L}_{\text{k}}^{L2R}(\theta),
\end{equation}

\begin{equation}
\text{s.t.   } \quad \mathcal{L}_{\text{k}}^{L2R} = \frac{1}{n_k} \sum_{n=1}^{n_k} \lVert z_{k}^{(n)} \rVert_2^2,
\end{equation}
where \(z_k^{(n)}\) represents the model representation of a specific layer for the client \(k\) and data point \(n\), and $\alpha^{L2R}$ is a hyper-parameter. It was demonstrated that in domain generalization tasks, FedL2R surpasses relevant FL baselines, such as Federated Adversarial Domain Generalization (FedADG)~\cite{zhang2021federated}.

The aforementioned works mainly targeted cross-device decentralized learning. Yet, existing research has underscored the value of cross-silo FL, where fewer but resource-rich organizations like manufacturers, hospitals, and banks serve as clients~\cite{kairouz2021advances}.
FL has recently received attention in the manufacturing sector~\cite{berghout2022federated}. For instance, Ahn et al.~\cite{ahn2023federated} employed FL for anomaly detection and predictive maintenance in pumps. FL methods were developed for defect detection in metal AM~\cite{mehta2022federated} and fault diagnosis in rotating machinery~\cite{mehta2023federated}. Despite these advances, there is a notable lack of research on FL for UMW. CM in UMW poses more significant challenges, due to complex process physics and limited understanding of process mechanisms. Moreover, the presence of multiple types of anomalies complicates online monitoring~\cite{lu2023online}.

While FL can address the data availability issue in CM applications by enabling the secure collaboration of data sources, recent applications of FL in CM have shown that domain shift issues can adversely affect the model performance~\cite{berghout2022federated}. To overcome this challenge, several federated transfer learning and personalization techniques have been proposed, e.g., \cite{mansour2020three, kevin2021federated, chung2023federated}. Kevin et al.~\cite{kevin2021federated} developed a federated transfer learning method capable of cross-domain knowledge transfer. While this method achieved better accuracy in the target application with less data and time, the framework was not tested on manufacturing datasets. Another critical limitation of this study is the lack of test data from entirely unseen configurations or domains. In a separate effort, Chung et al.~\cite{chung2023federated} proposed the FedEntropy framework, a personalized FL method using flat minima, which is known to perform better on unseen data than sharp minima, with the goal of enhancing generalization. Yet, the effectiveness of their framework in real-world scenarios remains unknown, as it was evaluated using a synthetic dataset on aircraft engine degradation.

In summary, although there are studies addressing the issue of domain shift within the FL regime, the area of domain generalization has still been limitedly studied. Moreover, while FL has been investigated in manufacturing applications, methods that specifically address domain shift, particularly domain generalization problems, in these applications, including CM for UMW, are generally lacking.

\section{Problem Formulation and Data Description}\label{sec:problem}
\subsection{Problem statement}

 This paper investigates the domain shift problem, particularly domain generalization, in CM for UMW, under the following general assumptions.

\begin{itemize}
    \item Data privacy preservation: The raw data remains localized and is not shared between clients and the central server.
    \item Similar feature space across datasets: We assume the availability of at least two datasets, each possessing a similar feature space but potentially encompassing distinct domain spaces and classification tasks.
    \item Data distribution heterogeneity: It is assumed that participating domains have distinct but related data distributions. This diversity motivates the need for domain generalization. 
    \item Common task objective: Each dataset's domains share a common task or objective despite their differences, and the goal is to develop a model that can generalize across these domains to make accurate predictions on unseen data.
    \item Evaluation on unseen domains: The model's performance is evaluated based on its ability to generalize to previously unseen domains during testing.

\end{itemize}
Consider a scenario where each dataset consists of \(k\) domains, denoted as \(D_{\text{1}}\), \(D_{\text{2}}\), \ldots, \(D_{\text{k}}\), serving as source domains, while \(D_{\text{target}}\) represents the target domain. The primary goal in addressing domain shift is to develop a model that can effectively perform on the target domain, utilizing data from the source domains. We assume that each client in the FL network corresponds to a company with access solely to data from a specific domain. Consequently, clients are limited to data points from their respective source domains. This scenario creates a significant challenge for model generalization, as it lacks the benefit of simultaneous access to data from multiple domains. Addressing this challenge requires innovative approaches within the FL paradigm.

\subsection{Dataset description and preprocessing}

We utilize datasets from a previous work~\cite{meng2023explainable} for CM of UMW. In UMW, the quality of joint is influenced by the interplay of process configurations, tool conditions, and workpiece properties~\cite{lu2023online}. The experiments were carried out using the Branson Ultraweld L20 Ultrasonic welding machine that is equipped with an online monitoring system. This monitoring system collects data at sampling rates of over 200 kHz from four sensors: a built-in linear variable differential transformer (LVDT), a built-in power sensor, an acoustic emission (AE) sensor, and a microphone.

\begin{table}[htbp]
\centering
\caption{Design of experiments for domain groups M, S, and T configurations~\cite{meng2023explainable}.}
\label{tab:combined_experiments}

\begin{subtable}{\textwidth}
\centering
\caption{Domain group M configuration}
\label{tab:experiment_design}
\begin{tabular}{@{}ccccc@{}}
\toprule
Domain & Material & Welding time & Data size & Classification goal \\ 
\midrule
AC     & Al-Cu    & 0.5 s        & 200          &  \\
CC     & Cu-Cu    & 0.9 s        & 200          & TC1, TC2, TC3, TC4 \\
CA     & Cu-Al    & 0.5 s        & 200          &  \\
AA     & Al-Al    & 0.9 s        & 200          &  \\
\bottomrule
\end{tabular}
\end{subtable}

\vspace{1em} % Add some vertical space between the tables

\begin{subtable}[t]{0.5\textwidth}
\centering
\caption{Domain group S configuration}
\label{tab:domain_group_s}
\resizebox{\textwidth}{!}{% Resize table to fit within subtable width
\begin{tabular}{@{}ccc@{}}
\toprule
Domain & Data size & Classification goal\\ 
\midrule
Clean    & 90 &  \\
Polished & 90 & New, Worn, DMGD  \\
Contam   & 90 &  \\
\bottomrule
\end{tabular}
}
\end{subtable}%
\begin{subtable}[t]{0.5\textwidth}
\centering
\caption{Domain group T configuration}
\label{tab:domain_group_t}
\resizebox{\textwidth}{!}{% Resize table to fit within subtable width
\begin{tabular}{@{}ccc@{}}
\toprule
Domain & Data size & \vspace{1mm}Classification goal \\ 
\midrule
DMGD & 90 &  \\
New  & 90 & Clean, Polished, Contam\vspace{1.85mm}\\
Worn & 90 &  \\
\bottomrule
\end{tabular}
}
\end{subtable}
\end{table}

In these datasets, different types of materials, tool conditions, and surface conditions are categorized into the following three distinct groups.

Domain group M: In this group, the welding samples are generated by combining different materials from either 0.25 mm thick Copper (Cu) or Aluminum (Al) sheets, each subjected to varying welding times, as detailed in Table~\ref{tab:experiment_design}. Each domain within this group comprises 200 samples, evenly distributed across four tool conditions: new horn/new anvil (TC1), new horn/worn anvil (TC2), worn horn/new anvil (TC3), and worn horn/worn anvil (TC4).

Domain group S: In this setup, 0.20 mm Cu sheets are welded for fixed welding
time of 1.0 second. As detailed in Table~\ref{tab:domain_group_s}, different domains are characterized by distinct surface conditions, including a ``Clean'' surface prepared using alcoholic wipes, a ``Polished'' surface achieved by sandpapering the contact faces, and a ``Contam'' (Contaminated) surface created by contaminating the surface by cutting fluid. 30 samples were generated for each combination of workpiece surface condition and tool condition, resulting in a total of 270 samples. The objective of the S domains is to classify three tool conditions: new horn/new anvil (New), worn horn/worn anvil (Worn), and damaged horn/damaged anvil (DMGD).

Domain group T: In this domain group, the data is identical to that in domain group S. However, the primary objective shifts to classifying the surface conditions while maintaining different domains representing distinct tool conditions, as outlined in Table~\ref{tab:domain_group_t}.

To tackle the challenge of high dimensionality, we choose discrete wavelet transformation (DWT) for its demonstrated effectiveness in prior research~\cite{meng2022physics}. This approach yields a total of 624 features for each data instance, significantly reducing dimensionality while retaining essential information.

\section{Proposed Methodology}\label{sec:method}
This section presents the details of the proposed FTL-TP framework. Specifically, Section~\ref{sec:framework} elaborates on the novel approach used for aggregating model parameters. This aggregation method, along with the loss function introduced in Section~\ref{sec:loss}, aims to overcome the domain shift issue in CM and enable domain generalization capabilities.

\subsection{The FTL-TP framework}\label{sec:framework}
In this framework, knowledge is transferred between at least two domain groups that have the same feature space. In a typical NN-based model, different features are extracted in different layers.
The lower layers extract low-level features that are very likely to
be transferrable between heterogeneous applications~\cite{kevin2021federated}. Therefore, the goal is to train the initial layers across various domain groups commonly, to facilitate the collaborative extraction of low-level information of the common feature space. This approach not only broadens the data pool for these shared layers but also enables them to see data from different domain spaces. Additionally, the framework is designed to yield a personalized final model for each domain group, to customize these models to their specific tasks. To achieve this personalization, the framework leverages data from clients within the same domain group to train a set of upper layers. These layers are then integrated with the earlier-mentioned lower layers to construct a comprehensive NN model. This strategy of knowledge transfer and model personalization is anticipated to significantly improve the performance of the final models, especially in tasks requiring domain generalization.

\begin{figure}[h]
\centering
\includegraphics[width=0.9\linewidth]{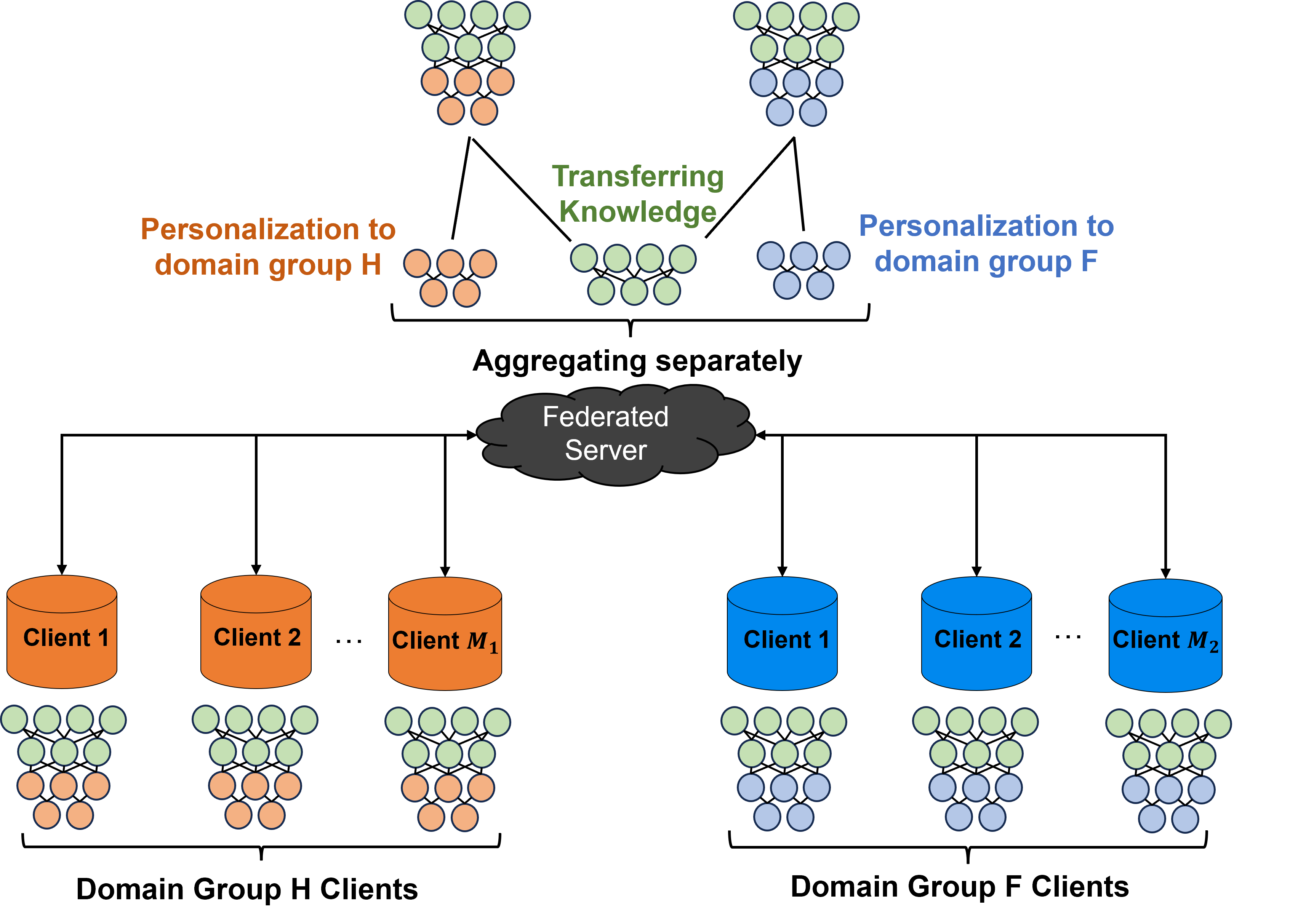}
\caption{Illustration of the proposed FTL-TP structure. Note that in this framework, the number of classes for each domain group and the neuron count in the blue and orange layers can differ.
}
\label{fig:structure}
\end{figure}

Utilizing a shared base layer among all clients, accompanied by personalized layers for each, has been previously implemented in the FedPer algorithm~\cite{arivazhagan2019federated}. However, FedPer restricts collaboration among clients with similar tasks, as it does not allow for the sharing of personalized layers with the server. In contrast, our approach promotes collaboration by enabling the separate aggregation of common personalized layers among clients with identical tasks and domain groups. 

For simplicity, we illustrate our framework using two CM classification tasks with domain shift issues, though it is designed to be scalable for more tasks. Figure~\ref{fig:structure} presents the details of implementing the FTL-TP framework on two hypothetical datasets from domain group H and domain group F, each associated with a specific task and assigned to a separate group of clients. In the first step, the server initializes and distributes two distinct models to the two groups of clients. Each model consists of shared base layers, depicted in green in the accompanying figure, followed by personalized layers tailored specifically to each dataset. For domain group H, these personalized layers are highlighted in orange, while for domain group F, they are shown in blue.

In the next step, each client trains the provided model on its local data and returns the updated model to the server. Then the server aggregates the base layers from all clients while only aggregating the subsequent personalized layers from the clients of the same domain group. The server then sends the two aggregated models back to the clients. These steps iterate until both models converge. This approach ensures that the base layer, responsible for learning a general data representation, benefits from both datasets, while the subsequent layers, which extract task-specific information, are learned exclusively from their respective datasets.

\begin{figure}[H]
  \centering
  \begin{minipage}{\linewidth}
    \begin{algorithm}[H]
    \caption{FTL-TP Method }
    \label{alg:FTL_TP}
    \begin{algorithmic}[2]
      \State \textbf{Server side:} 
      \State Initialize parameters $\theta_B^{0}$, $ \theta_{P1}^{0}$, $ \theta_{P2}^{0}$
      
      \For{each server round $t = 1, 2, \ldots$}
        \State select random subsets $m_1 , m_2$ with [$C_1 \cdot M_1$] , [$C_2 \cdot M_2$] number of clients
        \State Send the model $\theta_B^{t-1}+\theta_{P1}^{t-1}$ and $\theta_B^{t-1}+\theta_{P2}^{t-1}$ to the selected $m_1$ and $m_2$ clients respectively. 
        \For{each client $k_1 \in m_1$ in parallel}
          \State $\theta_B^t (k1), \theta_{P1}^t(k1)\gets$ ClientUpdate($k_1, \theta_B^{t-1}, \theta_{P1}^{t-1}$)
        \EndFor
        \For{each client $k_2 \in m_2$ in parallel}
          \State $\theta_B^t(k2), \theta_{P2}^t(k2) \gets$ ClientUpdate($k_2, \theta_B^{t-1}, \theta_{P2}^{t-1}$)
        \EndFor
        \State $N_{P1}  \gets \sum_{k1=1}^{m_1} n_{k1}$ 
        \State $N_{P2}  \gets \sum_{k2=1}^{m_2} n_{k2}$ 
        \State $N \gets N_{P1} + N_{P2}$
        \State $\theta_B^{t} \gets \frac{1}{N} ( \sum_{k1=1}^{m1} {n_{k1}} \cdot \theta_B^t(k1) + \sum_{k2=1}^{m2} {n_{k2}} \cdot \theta_B^t(k2))$
        \State $\theta_{P1}^{t} \gets \frac{1}{{N_{P1}}} \sum_{k1=1}^{m1} {n_{k1}} \cdot \theta_{P1}^t(k1) $
        \State $\theta_{P2}^{t} \gets \frac{1}{{N_{P2}}} \sum_{k2=1}^{m2} {n_{k2}} \cdot \theta_{P2}^t(k2) $      \EndFor
    \end{algorithmic}

    \begin{algorithmic}[2]
    \State \textbf{Client side:} % Mention "Client-side" here
    \For{each local epoch $i = 1, 2, \ldots, E$}
      \For{each mini-batch $b$ of size $B$}
        \State $\theta \gets \theta - \eta \nabla \mathcal{F}(\theta, b)$
      \EndFor
    \EndFor
    \State Send the updated local model $\theta$ to the server
    \end{algorithmic}
    \end{algorithm}
  \end{minipage}
\end{figure}

In this approach, the number of base layers can be considered as a hyper-parameter ($K_B$). All parameters preceding this layer are considered base parameters and are represented by $\theta_B$. Furthermore, $K_{P1}$ and $K_{P2}$ represent the number of personalized layers in each model, with their corresponding parameters denoted as $\theta_{P1}$ and $\theta_{P2}$. The total number of clients in each domain group is denoted by \( M_1 \) and \( M_2 \). Hyper-parameters, such as client fractions ($C_1$, $C_2$), must be specified individually for each client set in the FTL-TP setup. Notably, as the same base layer is employed for both sets of clients, a single learning rate parameter ($\eta$), client epoch ($E$), and batch size ($B$) are defined to enhance convergence. Algorithm~\ref{alg:FTL_TP} shows the details of the server and client steps.

\subsection{The FTL-TP loss function}\label{sec:loss}

Using a shared base layer in models that serve clients with different domains and tasks, may lead to potential divergence issues. To counter this, we incorporate the FedProx~\cite{li2020federated} loss term into the loss function of each local client. This addition effectively reduces the disparity in local updates. Moreover, to address domain shift challenges, we include an L2R term~\cite{nguyen2022fedsr} on the output of the shared base layer(s). The L2R term aids in learning a simpler representation from the feature space by constraining this representation, thus improving the model's capacity to effectively generalize across various domains. Consequently, the objective function of our framework is defined as:
\begin{equation}
\min_{\theta} F_k(\theta) + \alpha^{L2R} \cdot \mathcal{L}_{\text{k}}^{L2R}(\theta) + \frac{\mu}{2} \|\theta - \theta_t\|^2,
\end{equation}
\begin{equation}
\text{s.t.     }\mathcal{L}_{\text{k}}^{L2R} = \frac{1}{n_k} \sum_{n=1}^{n_K} \lVert z_{k}^{(n)} \rVert_2^2 ,
\end{equation}
where $\alpha^{L2R}$ and $\mu$ are hyper-parameters. Also, $z_{k}^{(n)}$ is the output of the latest base layer for client $k$ with data point $n$. By optimizing this function, we aim to not only achieve superior generalization but also to enhance the model's convergence

\subsection{NN architecture}\label{sec:network}   

Within the proposed NN architecture, the primary goal is to accurately classify each input signal into the correct category. The network, trained on source domains, is evaluated on unseen target domains to assess its performance.

624 features that are extracted using DWT are used as the input data for classification. An MLP architecture with four layers is designed. The model takes 624 DWT features as input and the output layer provides the classification results for each specific dataset (4 classes for domain group M and 3 classes for domain groups S and T). For the first three hidden layers, the model has 175, 125, and 50 neurons, respectively. The rectified linear unit (ReLU) is chosen as the activation function, and adaptive moment estimation (Adam) is selected as the solver.

\subsection{Edge-cloud implementation}\label{sec:edge}

To emulate a manufacturing environment,  models are trained on Raspberry Pis which are commonly used as edge devices. Typically, multiple welding machines operate within a single factory; thus, each Raspberry Pi can represent a factory, training models for up to four welding machines. These devices—emulating companies—connect to a remote server via the Internet for collaborative learning. The entire training process is coordinated by a remote server, which defines all learning strategies, including the methods used, model hyperparameters, the selection and number of participating clients, and the number of server rounds. Clients receive the model and hyperparameters from the server and conduct their training independently. 

The RabbitMQ broker, a publish-subscribe system, connects the server node and client nodes in a bi-directional manner \cite{dossot2014rabbitmq}. In RabbitMQ, messages are published to exchanges and delivered to queues based on routing rules defined by bindings and routing keys. This decoupling of producers and consumers enhances scalability, flexibility, and robustness. Secure connections to the RabbitMQ broker require the server and clients to authenticate using a username and password.

\begin{figure}[h]
    \centering
    \makebox[\textwidth][c]{\includegraphics[width=1.2\textwidth]{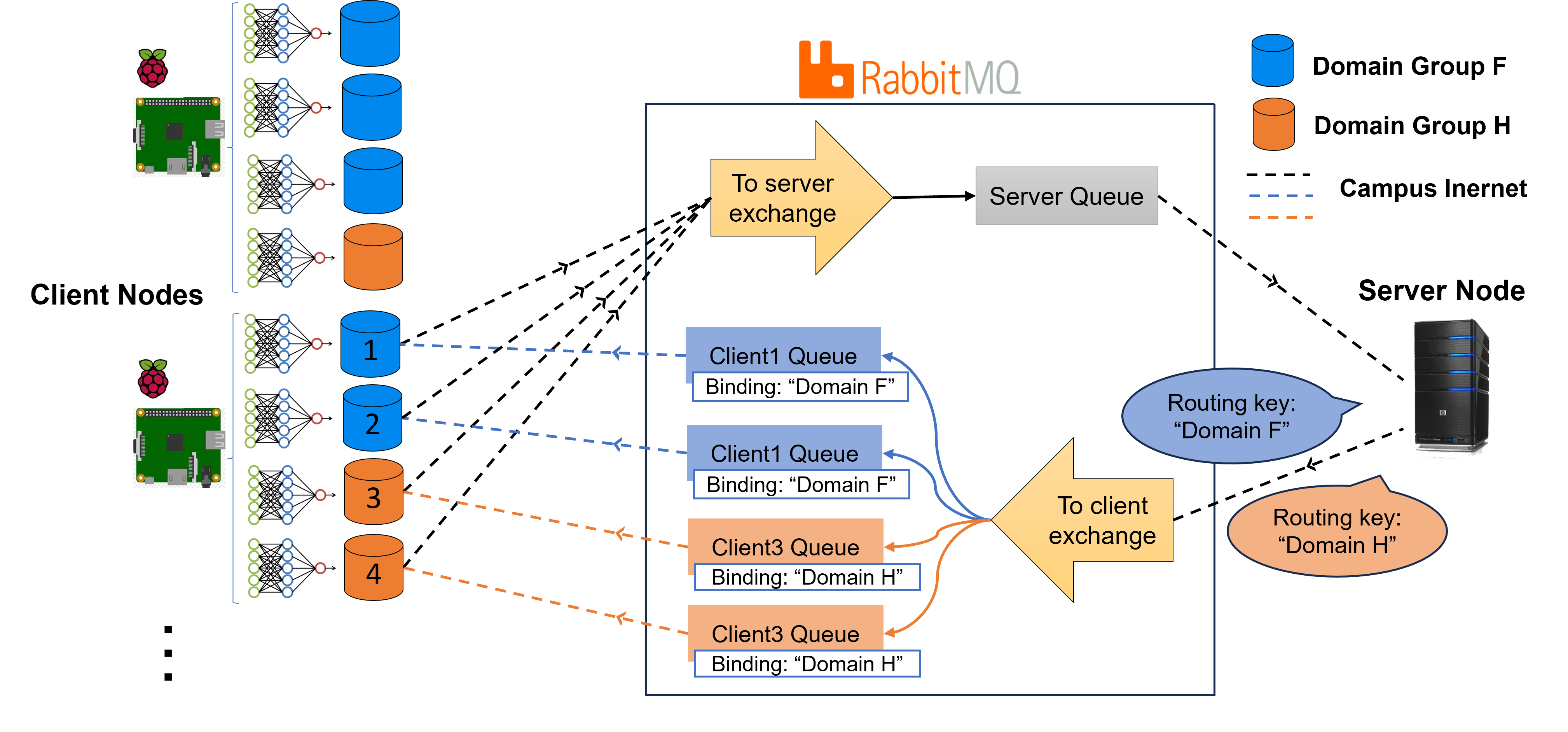}}
\caption{A schematic of edge-cloud implementation for the FTL-TP method.}
\label{fig:edge}
\end{figure}

Figure~\ref{fig:edge} illustrates the details of message routing in our implementation. For simplicity, we consider clients on only one Raspberry Pi—two from domain group H and the other two from domain group F. Initially, the server publishes two initialized models and corresponding hyperparameters as two JSON files each tagged with a specific routing key, ``Domain F''  or ``Domain H,'' to the broker. These routing keys are essential for the direct exchange model, labeled as ``To clients exchange,'' to route messages accurately to the corresponding client queues. Client1 and Client2 queues, bound with ``Domain F'', receive messages tagged with the ``Domain F'' routing key. Conversely, Client3 and Client4 queues, bound with ``Domain H'', are configured to receive messages intended for the domain group H. Then each client subscribes to its queue to retrieve its parameters for the next round. Following the training of the received models on their private data, each client publishes a JSON file with their updated parameters and a topic indicating their domain group to the ``To server exchange.'' This exchange is another direct model that forwards messages to the ``Server Queue.'' The server subscribes to this queue to retrieve all the updated models from all clients. Upon retrieving the messages, the server identifies their corresponding domain groups using their topics as filters. It then aggregates the weights as per the framework shown in Figure~\ref{fig:structure}, resulting in two updated models. These updated global models are then published to the broker as a JSON file to continue the process. Note that clients, the broker, and the server are located on separate nodes, which are connected through the Internet.

\section{Case Studies} \label{sec:case}

This section describes the experimental setup and compares the FTL-TP method with multiple baseline methods to demonstrate its effectiveness. Additionally, we investigate the performance of FTL-TP under unbalanced data distribution and different client fractions. The training efficiency of FTL-TP in an actual cloud manufacturing environment is also evaluated.

\subsection{Experimental setup}
 
Our setup includes four Raspberry Pi devices, each capable of running four clients, and a computer that manages and distributes messages using a RabbitMQ broker. The computer is equipped with a 12th Gen Intel(R) Core(TM) i9-12900F processor, an NVIDIA GeForce RTX 3070 Ti 8GB GDDR6X, and 16 GB of RAM. Each Raspberry Pi 4B model used is powered by a 4-core ARMv8 CPU and is equipped with 4 GB of RAM. For network connectivity, we utilize the Internet network provided by the University of Illinois. 
For software, we use TensorFlow (2.10.0), Python (3.9.2), and the Raspberry Pi OS (64-bit) based on Debian Bullseye. We use the RabbitMQ broker with the Pika (1.3.2) library in Python for message routing between the server and clients.

For training, we use 5-fold cross-validation for all models. Instead of random initialization, the server uses ``Kaiming normal'' to initialize the weights of all the layers. All models use categorical cross-entropy as the base loss function. As explained in Section~\ref{sec:problem}, three datasets with domain groups M, S, and T are available. For simplicity, the proposed method will concentrate on the collaboration between just two datasets. The aim is to investigate any two out of three combinations of these datasets. However, we do not combine domain groups S and T, as they represent the same data signals but with different domain shifts and classes. Therefore, the focus is on the combinations of domain groups M vs S, and M vs T.

The client distribution is as follows: One domain of each domain group is considered as the target domain, and each of the other domains is equally distributed as the source domains among three clients at random. This means each client has access to data from only one domain, which is similar to real-world company scenarios. This distribution results in 9 training clients for domain group M and 6 each for domain groups S and T. As depicted in Tables~\ref{tab:material_domain}, \ref{tab:surface_domain_distribution}, and \ref{tab:tool_domain_distribution} in \ref{appendix:first}, clients in domain group M have about 67 data points each, while those in groups S and T have 30 data points each. The challenge in learning models suitable for domain generalization is heightened by the limited number of data points available to each client and the restriction of having access to only one domain's data.

In all experiments, the same NN model presented in Section~\ref{sec:network} serves as the basis for all training methods (FTL-TP, IL, CL, CTL, and other FL paradigms). We report the accuracy of the proposed method for each domain, treating it as the target for one dataset and considering all possible target domain combinations for the other dataset. For instance, in the scenario of training domain group M vs domain group T, to report the accuracy for the Al-Al target in the M domain group, we conduct the experiment three times. Each time involves selecting a different domain from domain group T as the target domain. This approach ensures a thorough evaluation of the FTL-TP method across a range of domain configurations, providing a detailed understanding of its performance in varying settings. 

To address the randomness in NN outcomes, each experiment is conducted five times. Consequently, for the Al-Al target in domain group M, each of the three target domain combinations from domain group T is tested five times, resulting in a total of 15 accuracy measurements. The reported accuracy represents the average and standard deviation of these tests. This is achieved by first calculating the mean and standard deviation for each of the three combinations separately, and then averaging these means and standard deviations across all three combinations. This process ensures a comprehensive and accurate reflection of the model's performance for the never-before-seen target. 

Through fine-tuning, the hyper-parameters for the FTL-TP method are set identically across both domain group combinations. The first layer functions as the base layer, with the subsequent three layers designated as personalized layers. The minibatch size is set to 8, and the learning rate to 0.0005, with one epoch selected for local training. Additionally, the proximal term and L2 regularization hyperparameters are determined to be 0.01 and 0.001, respectively. Experiments indicate that convergence for all clients is achieved before 150 server rounds, so this number of server rounds is used for all experiments with the proposed methods. Details on the hyperparameters for other models will be provided in their respective sections.

\subsection{Performance comparison with other learning paradigms} \label{sub:case_learning_paradigms}

\begin{figure}[htbp]
    \centering

    \begin{subfigure}[b]{0.8\textwidth}
        \includegraphics[width=\textwidth]{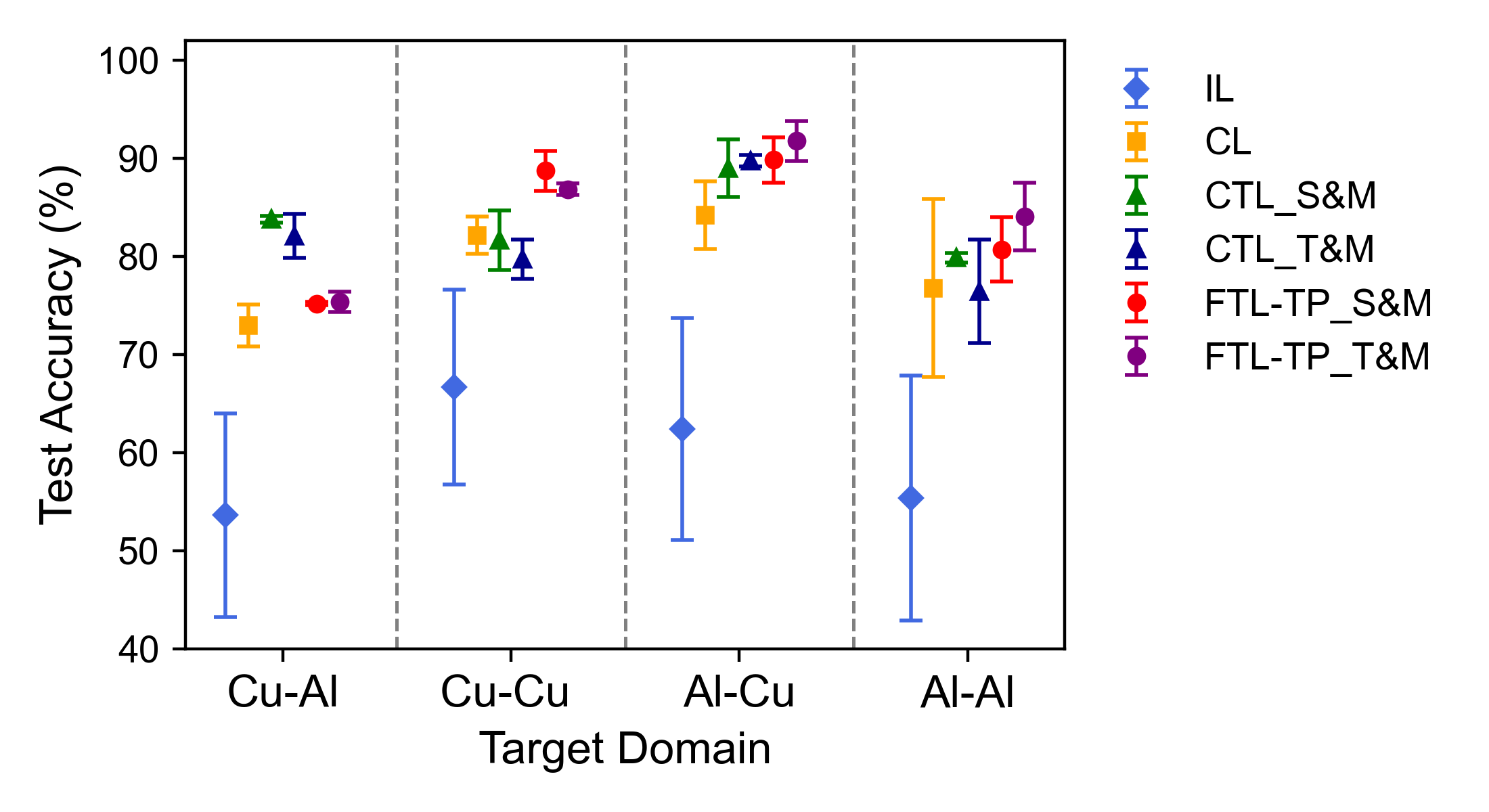}
        \caption{Tool classification accuracy for different material targets}
        \label{fig:IL-material}
    \end{subfigure}
    \begin{subfigure}[b]{0.8\textwidth}
        \includegraphics[width=\textwidth]{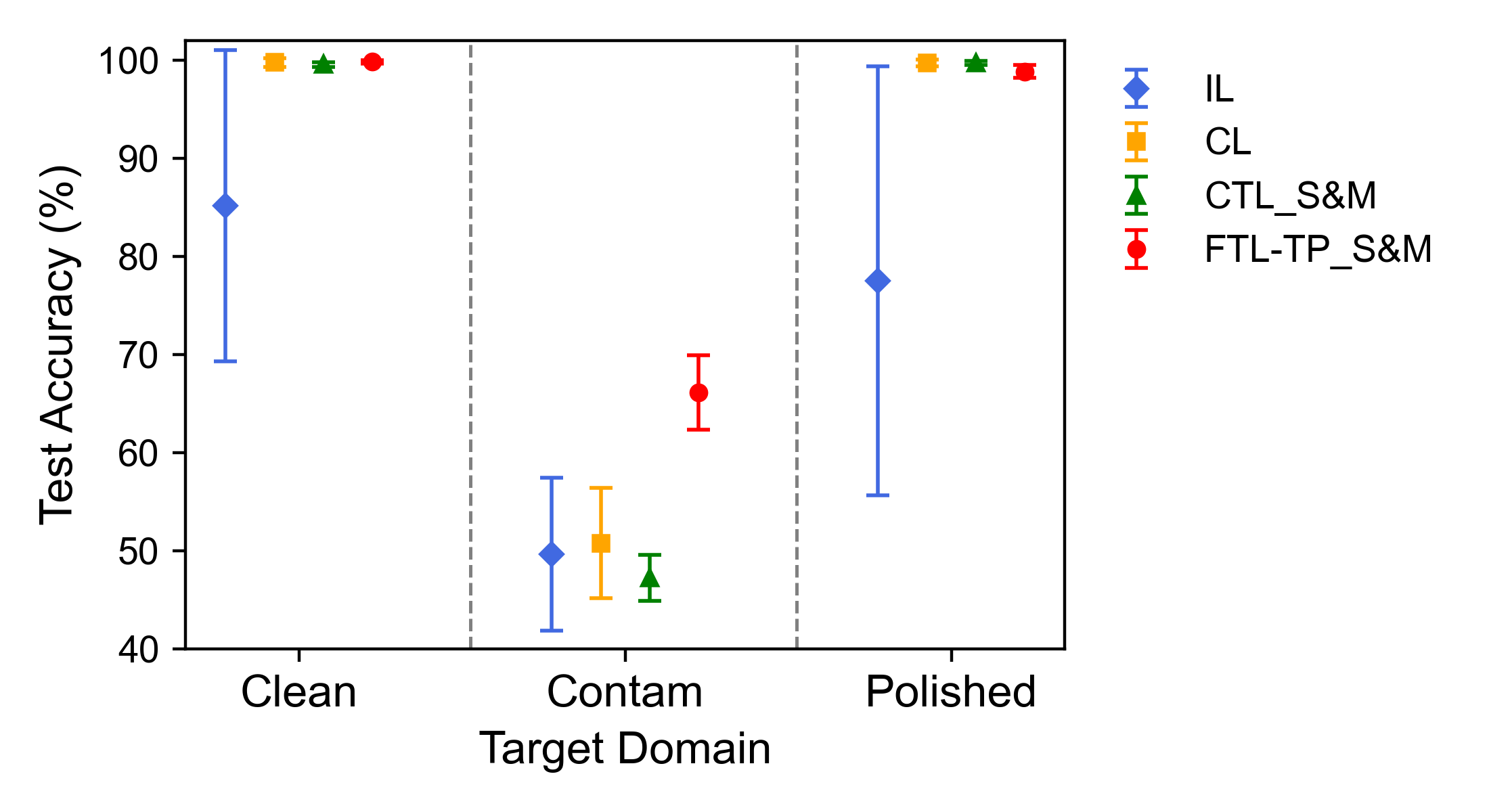}
        \caption{Tool classification accuracy for different surface targets}
        \label{fig:IL-surface}
    \end{subfigure}
    \
    \begin{subfigure}[b]{0.8\textwidth}
        \includegraphics[width=\textwidth]{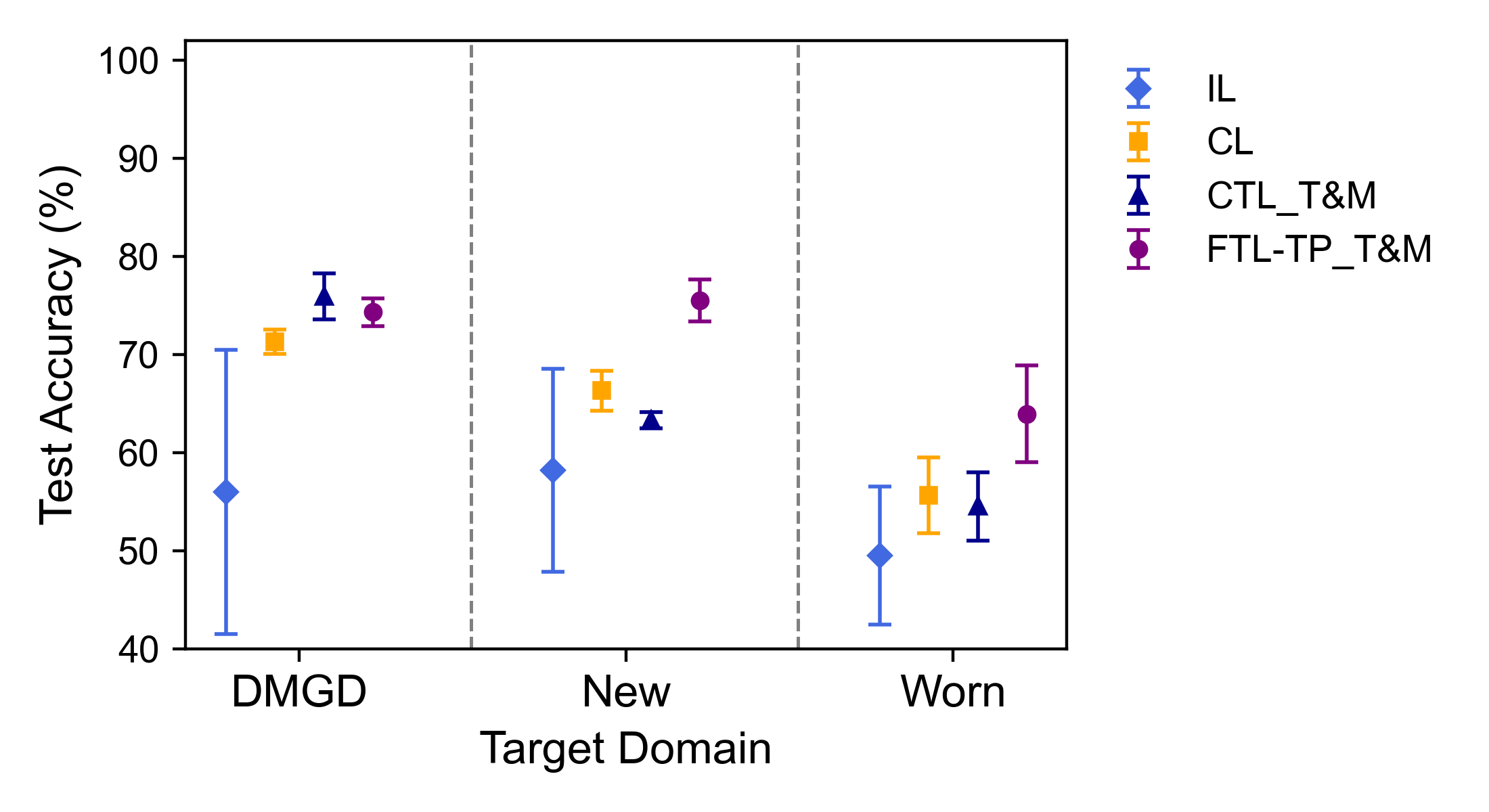}
        \caption{Surface classification accuracy for different tool targets}
        \label{fig:IL_tool}
    \end{subfigure}

    \caption{Performance comparison with IL, CL, and CTL. `S\&M' and `T\&M' refer to the domain group combinations `Surface vs Material` and `Tool vs. Material,' respectively.}
    \label{fig:IL_CL}
\end{figure}

This section aims to assess the domain generalization capability of the FTL-TP model in CM tasks for UMW. We compare FTL-TP with CL, individual learning (IL), and centralized transfer learning (CTL). In CL, data from all clients is combined for training. In IL, each client uses only their own data for model training. Given that FTL-TP involves knowledge transfer between datasets, we also study CTL, where the model is initially trained on one dataset and, after freezing the base layers (first layer in this case), it is fine-tuned with another dataset. To assess the domain generalization of the FTL-TP model, we evaluate it on data from unseen domains.

Figure~\ref{fig:IL_CL} shows the comparative results. In this study, the average performance of all clients trained individually is used to indicate the performance of IL. All experiments for both IL and CL were repeated five times. To ensure consistency, the same target data were employed across all models: IL, CL, CTL, and FTL-TP. After fine-tuning, hyperparameters are established for the IL, CL, and CTL models. These models are configured with a batch size of 8 and a learning rate of 0.0005. Each model underwent training for a total of 150 epochs.

The results demonstrate that the FTL-TP method surpasses both IL and CL across nearly all instances, with an average improvement of 20.83\% and 5.44\%, respectively. This improvement can show the FTL-TP model's effectiveness in transferring knowledge among clients working on two distinct CM tasks. Notably, in most cases, our method exhibits less variation in performance, suggesting enhanced robustness compared to these two models. Furthermore, on average, the FTL-TP method outperforms the CTL model with an improvement of 3.84\%.

\subsection{Performance comparison with other FL methods} \label{sub:case_FL}

\begin{figure}[htbp]
    \centering

    \begin{subfigure}[b]{0.8\textwidth}
        \includegraphics[width=\textwidth]{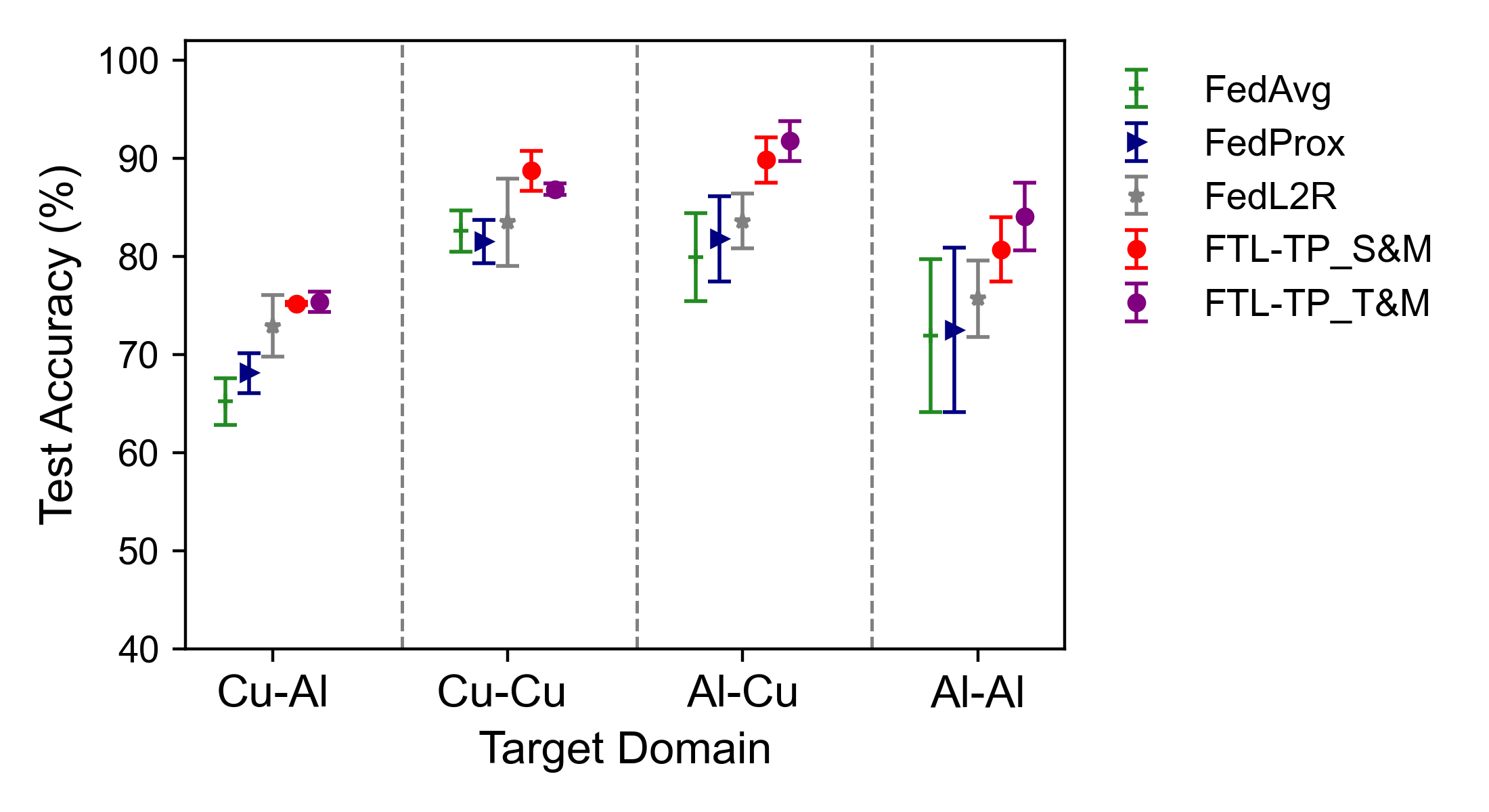}
        \caption{Tool classification accuracy for different material targets}
        \label{fig:FL-material}
    \end{subfigure}
    \begin{subfigure}[b]{0.8\textwidth}
        \includegraphics[width=\textwidth]{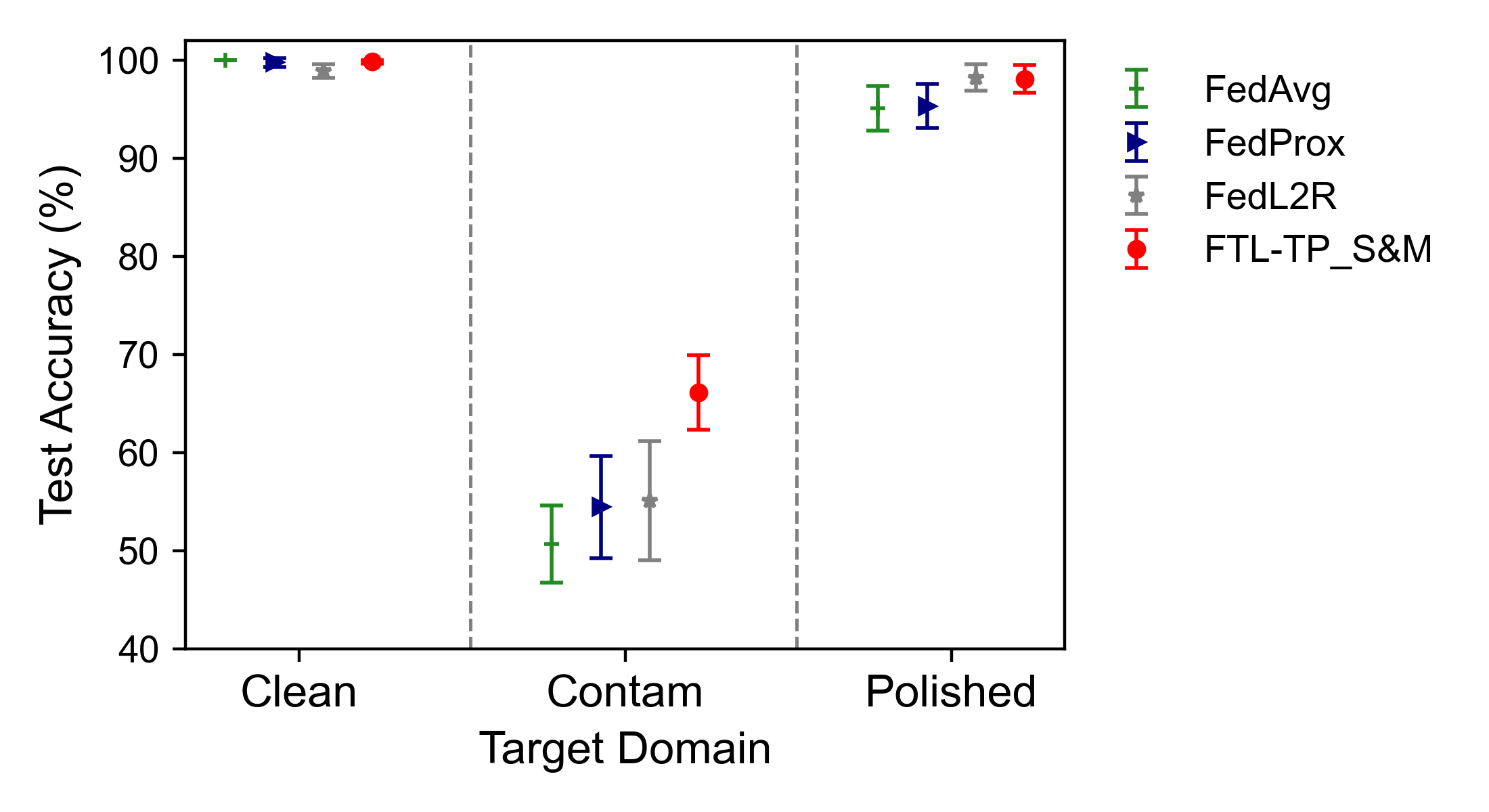}
        \caption{Tool classification accuracy for different surface targets}
        \label{fig:FL-surface}
    \end{subfigure}
    
    \begin{subfigure}[b]{0.8\textwidth}
        \includegraphics[width=\textwidth]{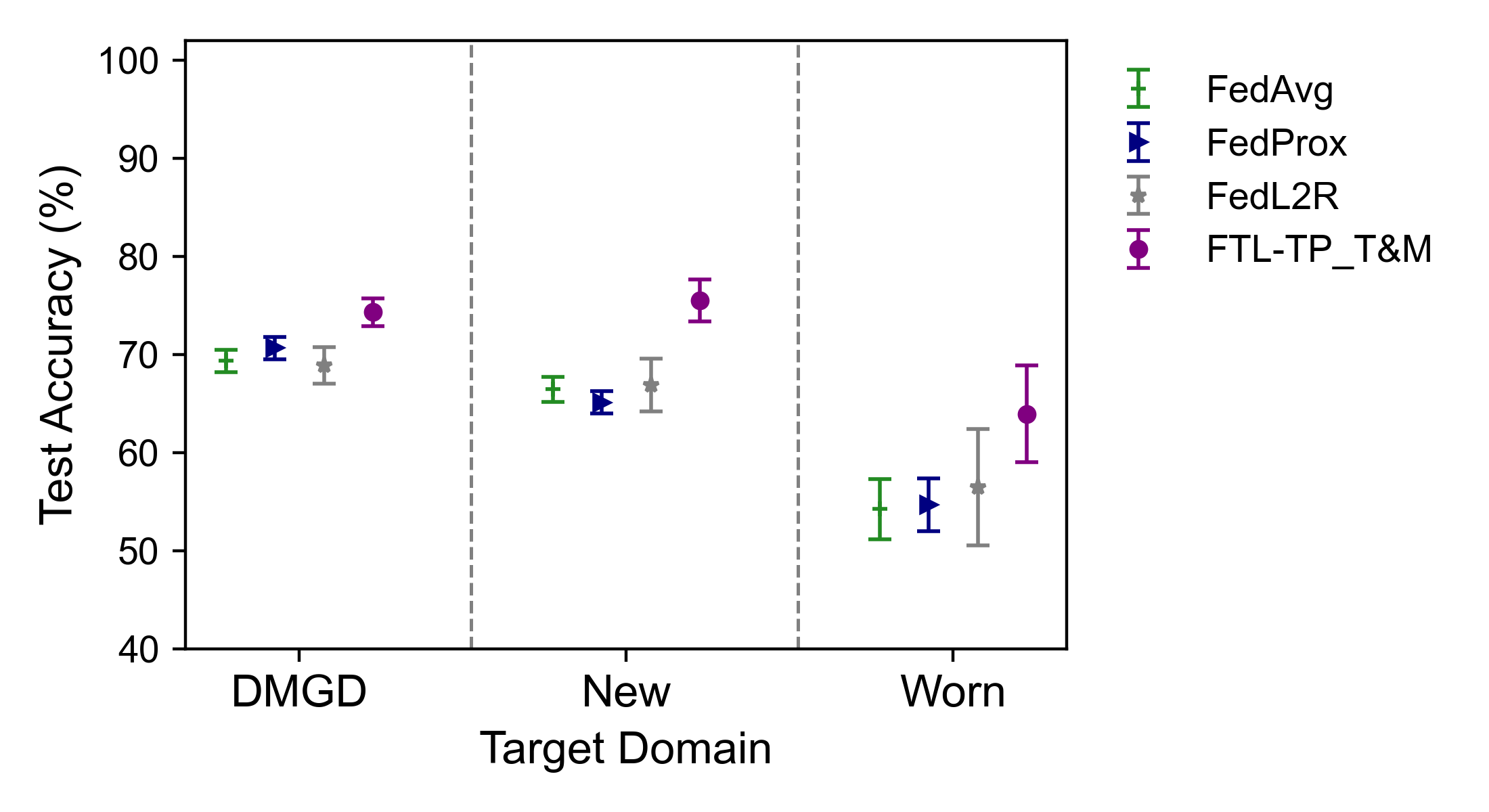}
        \caption{Surface classification accuracy for different tool targets}
        \label{fig:FL-tool}
    \end{subfigure}

    \caption{Performance comparison with FedAvg, FedProx, and FedL2R. `S\&M' and `T\&M' refer to the domain group combinations `Surface vs. Material' and `Tool vs. Material', respectively.}
    \label{fig:FL-Comp}
\end{figure}

This section compares FTL-TP with state-of-the-art FL models including FedAvg, FedProx, and FedL2R. These FL methods use the same client distribution as mentioned previously for a consistent comparison. To eliminate the randomness of NN models, we repeat all experiments five times. Following the fine-tuning process, a batch size of 8 and a learning rate of 0.0005 are chosen for these three models. Additionally, for FedProx, the proximal term is set at 0.1, and for FedL2R, the regularization term is 0.01. All these models iterate for 150 server rounds to ensure consistent and thorough training across all experiments.

Figure~\ref{fig:FL-Comp} presents the comparative results. It is seen that FTL-TP outperforms FedAvg, FedProx, and FedL2R with average improvements of 8.08\%, 7.19\%, and 5.35\%, respectively. It is worth noting that FTL-TP outperforms FedL2R, which is considered as a state-of-the-art domain generalization FL method, with a significant margin.

\subsection{Performance comparison between balanced and unbalanced data distribution}

\begin{figure}[h]
    \centering
    \begin{subfigure}[b]{.8\textwidth}
        \centering
        \includegraphics[width=\linewidth]{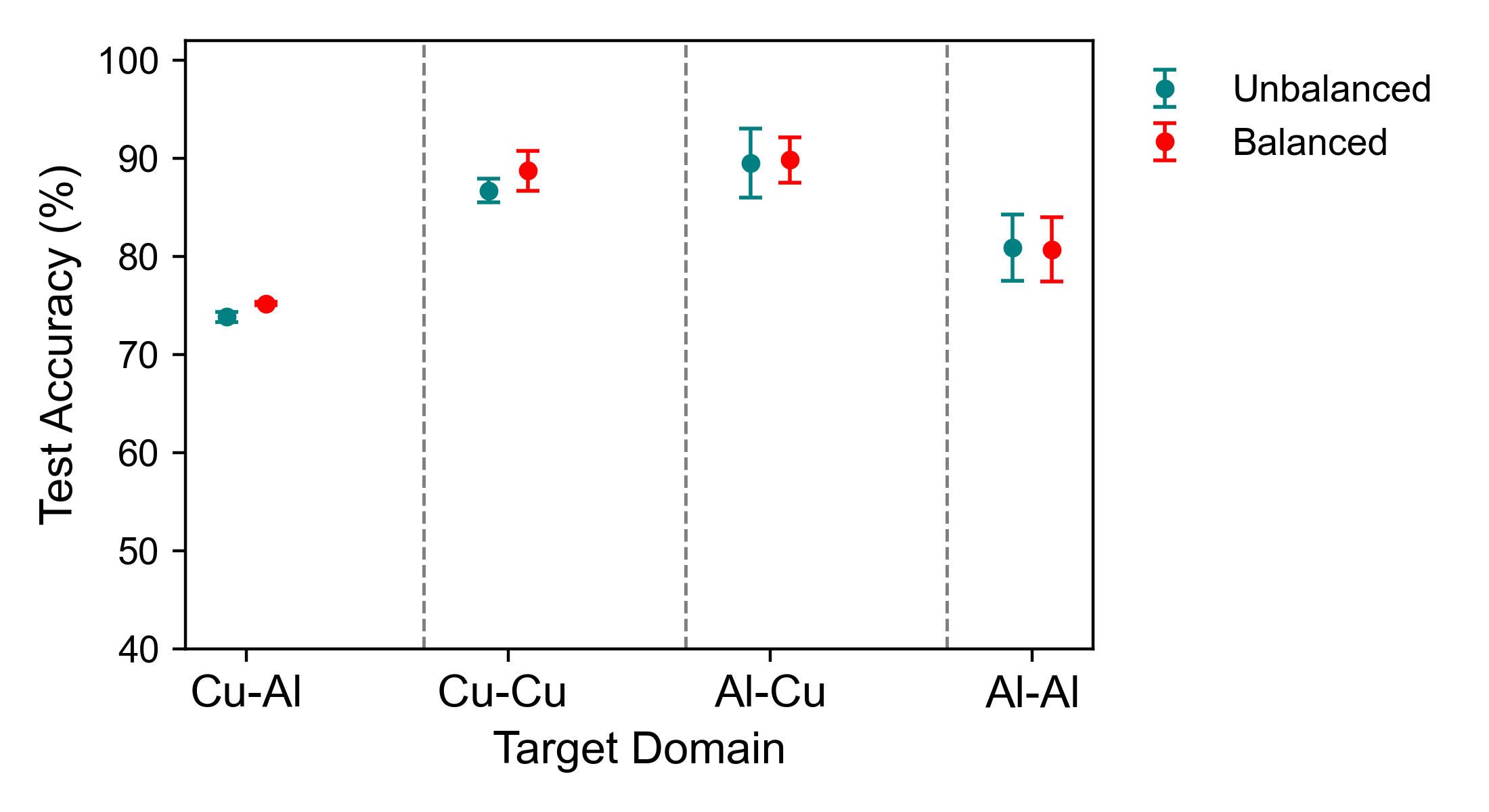}
        \caption{Tool classification accuracy for different material targets}
        \label{fig:unbalanced_material}
    \end{subfigure}
    % \hfill % This will add space between the two figures
    \begin{subfigure}[b]{.8\textwidth}
        \centering
        \includegraphics[width=\linewidth]{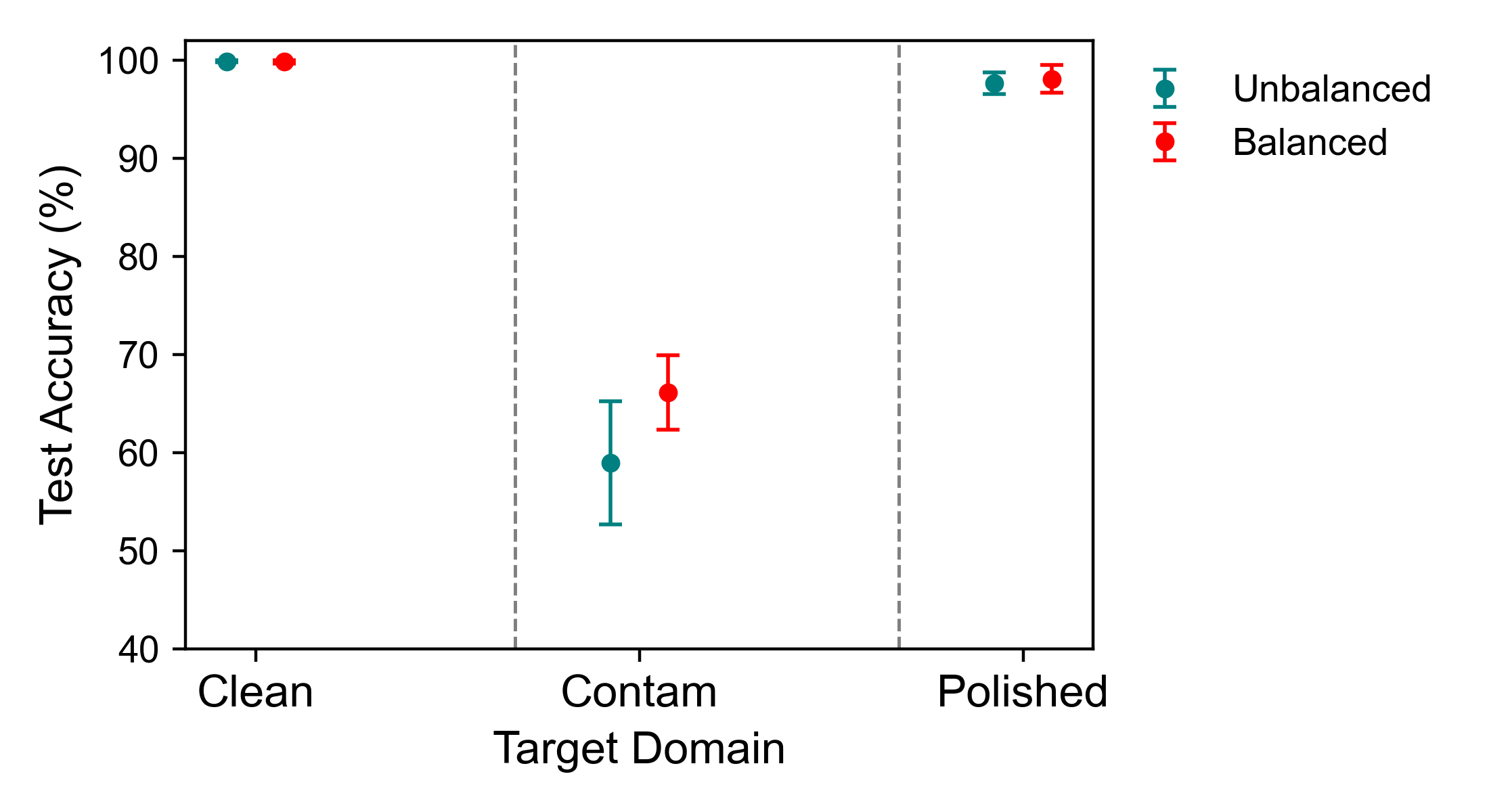}
        \caption{Surface classification accuracy for different tool targets}
        \label{fig:unbalanced_surface}
    \end{subfigure}
    \caption{Balanced vs unbalanced performance comparison for the FTL-TP model in domain groups S vs M combination.}
    \label{fig:balance_graphs}
\end{figure}

This section further mimics industrial scenarios by introducing unbalanced data distributions among clients, which is a significant challenge in FL performance. We investigate how the FTL-TP model copes with this challenge using domain groups M and S combination. To create an unbalanced distribution, we assign 30, 60, and 110 data points to the three clients of each training domain in domain group M, and 15, 25, and 50 data points to the clients in one training domain of domain group S. This aims to replicate a near-worst-case scenario in industrial settings.

Figure~\ref{fig:balance_graphs} shows the comparison between balanced and unbalanced data distributions for both domain groups M and S. Figure~\ref{fig:unbalanced_material} shows the accuracy of the FTL-TP model under unbalanced distribution is similar to that of the balanced distribution, with less than a 2.05\% drop, indicating robust performance for domain group M. However, Figure~\ref{fig:unbalanced_surface} shows that in domain group S, while the accuracy drops for the clean and polished domains as targets are negligible, the Contam domain exhibits a significant drop of 7.17\%. It is worth mentioning that, despite this significant drop, the performance of the FTL-TP method in the Contam domain under unbalanced distribution conditions still surpasses that of IL, CL, CTL, and other FL models under balanced conditions.

\subsection{Performance comparison for different client fractions} \label{sub:case_client_frac}

This section evaluates the data efficiency of FTL-TP. Specifically, the effect of varying client fractions on the training of the model is examined. For this purpose, we select a combination of domain groups M and S to conduct our investigation. For this analysis, in each server round, 1, 2, and all 3 of these clients are randomly selected. This results in collaborations involving 3, 6, and 9 clients from domain group M, as well as 2, 4, and 6 clients from domain group S to represent client fractions of 1/3, 2/3, and 1 respectively.

\begin{figure}[h]
    \centering
    \includegraphics[width=0.8\textwidth]{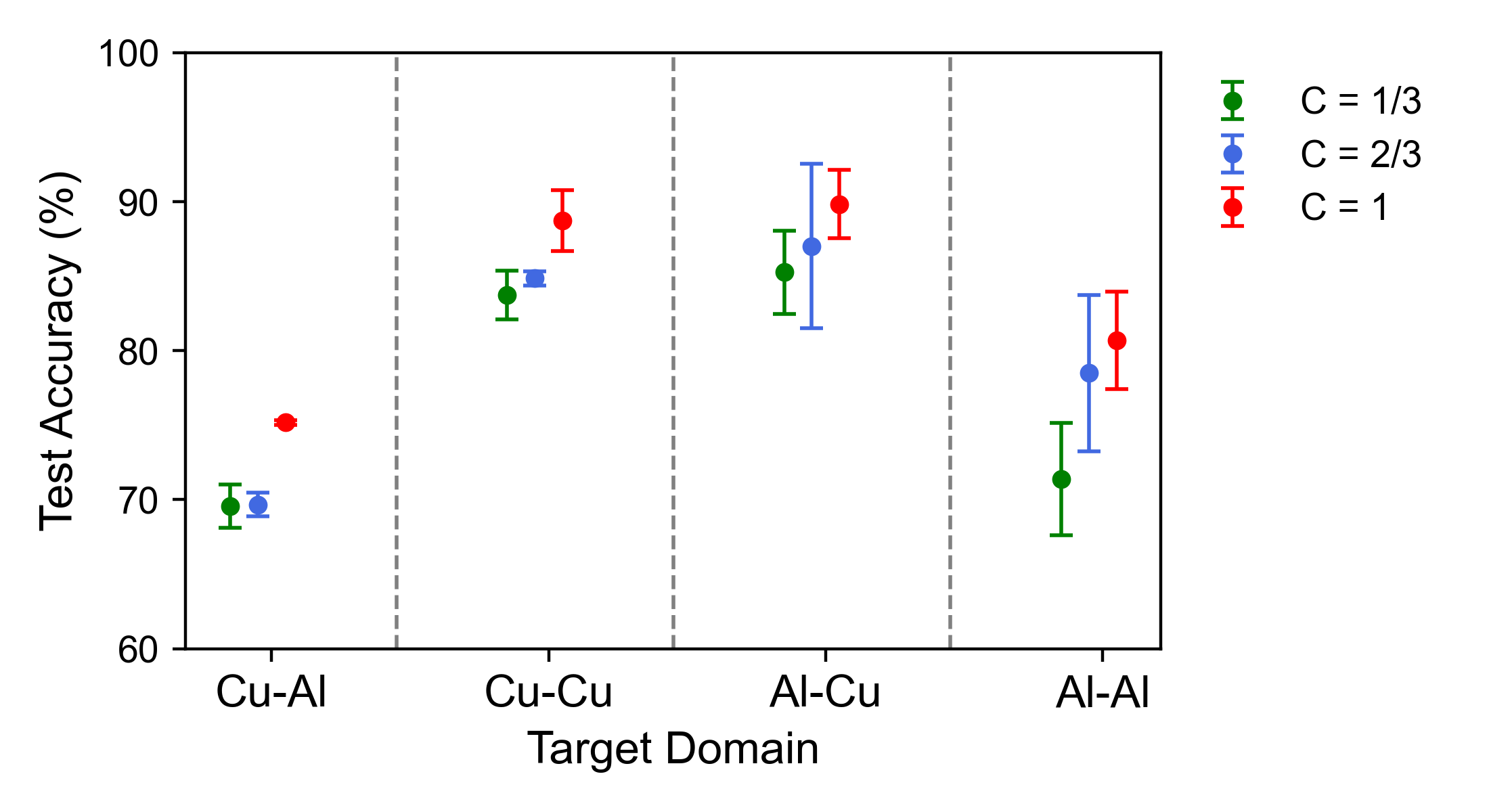}
    \caption{The FTL-TP's accuracy results for various material target domains in domain groups S vs M combination with different client fractions ($C$).}
    \label{fig:client_fraction}
\end{figure}

Figure~\ref{fig:client_fraction} compares FTL-TP's performance under different client factions. It is shown that within domain group M, the performance of the FTL-TP model enhances as the number of participating clients in the federation increases. This improvement is due to the larger and more diverse dataset available for training the model, consequently boosting the model performance. However, the performance of the FTL-TP method with a client fraction of 1/3 is comparable to that of other baseline FL methods with a client fraction of 1, and even the CL method.

\subsection{Comparative time analysis of FL methodologies} \label{sub:case_time}

Given that the FTL-TP method develops at least two personalized models during the training phase, evaluating its time efficiency is essential. We select the combination of domain groups M and S for this purpose. The aim is to compare the time efficiency of the FTL-TP model with that of single-dataset FL frameworks including FedAvg, FedProx, and a combination of FedL2R with FedProx, which employs a similar loss function to that of the FTL-TP method. 

We implemented the edge-cloud architecture presented in Section~\ref{sec:edge}. The time measurement was done from the moment the central server began distributing the initial weights to clients until the completion of 10 server rounds, with a client fraction of $C=1$. It is important to note that the proposed model outputs two distinct models, one each for domain groups M and S, whereas traditional single-dataset FL frameworks would output only one model for either domain groups M or S.

\begin{figure}[h]
    \centering
    \includegraphics[width=0.8\textwidth]{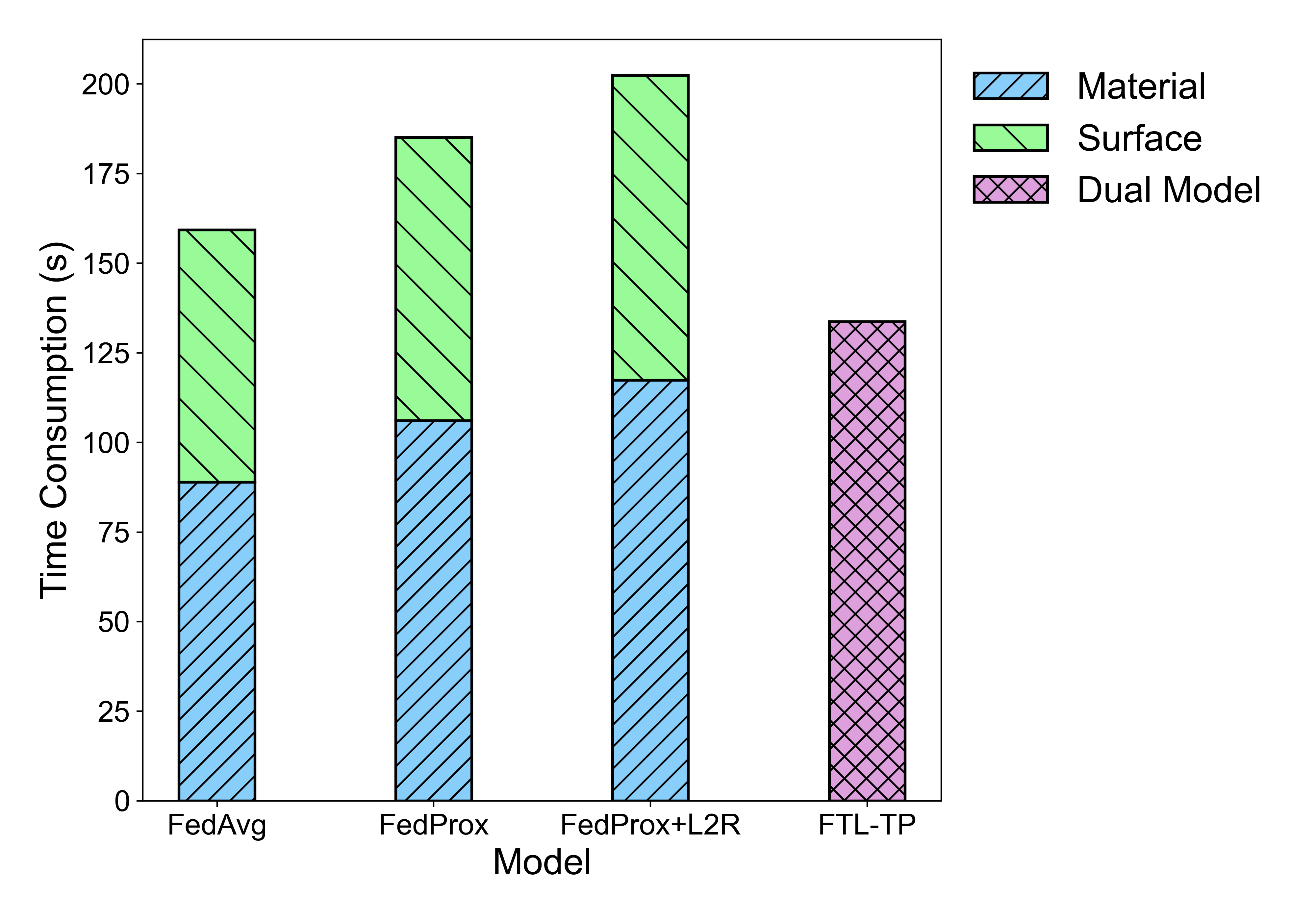}
    \caption{Comparison of time consumption for different FL models including FedAvg, FedProx, FedProx with L2 regularization, and the FTL-TP Model for training two NN-based models for domain group S (Surface) and M (Material).}
    \label{fig:time_consumption}
\end{figure} 

\ref{fig:time_consumption} compares the time consumption of different FL models. Specifically, we accumulate the total time required for two separate FL implementations to prepare two models and compare this with the time needed to produce the same two models using our proposed method in a single run. The results demonstrate that our proposed framework is less time-consuming compared to other FL approaches for achieving the same output.

\section{Discussion}\label{sec:discussion}

\subsection{CM performance}

The results reported in Sections \ref{sub:case_learning_paradigms} and \ref{sub:case_FL} lead to several important findings. First, individual clients do not have sufficient data to train their models effectively. This is clear from the significantly lower performance in CM for IL compared to CL, CTL, or any FL approach in this study. Therefore, it is beneficial for clients to participate in a collaborative model training paradigm. Second, the FTL-TP can solve domain shift issues in CM by collaboratively learning the specific characteristics of UMW datasets and using the proposed new loss function. Figures~\ref{fig:IL_CL} and \ref{fig:FL-Comp} show the FTL-TP's performance generally exceeds the performance of other FL paradigms, IL, and CL. This achievement underscores the benefits of learning a shared, simplified representation across various domain spaces. Third, a comparison between CTL and the proposed method, as shown in Figure~\ref{fig:IL_CL}, shows that the FTL-TP surpasses CTL in most cases. This observation reveals that the enhancement in domain generalization by the proposed method is not solely due to the transfer of knowledge between datasets. The manner of this knowledge transfer and the application of generalization techniques, such as the L2R loss, play crucial roles as well.

Additionally, we evaluate the proposed method under unbalanced data distribution, which is a key challenge in FL settings. Figure~\ref{fig:balance_graphs} shows that FTL-TP yields similar results in both balanced and unbalanced data conditions in most cases. The observed decrease in accuracy for the Contam domain as the target in Figure~\ref{fig:unbalanced_surface}, under unbalanced conditions, might be attributed to the specific characteristics of the contaminated surface condition, such as the presence of oil drops, which introduce a wide range of variability. This agrees with the findings from previous studies~\cite{meng2023explainable}, where CM tasks under contaminated surface conditions were shown to be more challenging. This decrease in accuracy could also be due to the model's increased sensitivity under unbalanced conditions. Moreover, the limited number of data points available to some clients in domain group S, potentially as few as 15, may further affect the model's accuracy. However, it is important to highlight that the unbalanced accuracy for the Contam domain in domain group S still exceeds the performance of IL, CL, CTL, and other FL models under balanced conditions, as evidenced by comparing Figure~\ref{fig:unbalanced_surface} with Figures~\ref{fig:IL-surface} and \ref{fig:FL-surface}.

\subsection{Training efficiency}

The efficiency of the proposed model is evaluated in terms of data and time efficiency. Focusing on data efficiency, Figure~\ref{fig:client_fraction} demonstrates that engaging more clients leads to improved performance. This improvement is linked to the FL model being trained on a larger dataset at each server round. A comparison of model performance with one-third of clients in this figure with Figure~\ref{fig:IL_CL} reveals that even limited collaboration among clients (just two clients for domain group S and three for domain group M) significantly boosts the accuracy of models over IL. Moreover, this minimum of collaboration yields results that are comparable to those achieved by CL.

Additionally, when comparing Figures~\ref{fig:FL-Comp} and \ref{fig:client_fraction}, it is evident that the outcomes of minimal collaboration in our proposed framework, using a client fraction of 1/3, are either comparable to or, in some instances, exceed the results of other FL approaches with a client fraction of 1. This finding suggests that the FTL-TP method, by including only two clients from domain group S (totaling 60 data points) and three clients from domain group M (totaling 200 data points), can outperform state-of-the-art FL methods that use all available data points from domain group M (600 data points). This notable improvement is likely due to the efficient representation learning that takes place between clients across both datasets.

Figure~\ref{fig:time_consumption} shows that our framework is more time-efficient in training models compared to traditional FL frameworks. For instance, the FedAvg framework, which is considered as the simplest FL model, requires over 20\% more time for training. When using an FL framework with the same loss function as FTL-TP, the extra time needed increases to more than 50\%. Notably, within the single-dataset frameworks for domain groups M and S, there are 9 and 6 clients, running on 3 and 2 Raspberry Pis, respectively. The FTL-TP method which is a dual-dataset framework includes 15 clients implemented on 4 Raspberry Pis, which can raise the latency in the edge-cloud implementation. Despite this, the proposed approach proves to be more time-efficient overall. Additionally, in the system developed with Raspberry Pis and a message broker for weight transmission, the training time for 10 server rounds is under two minutes, making FTL-TP suitable for real-world applications.

Based on these observations, it is evident that the FTL-TP framework is not only data-efficient but also exhibits considerable time efficiency. This combination of efficiencies makes this framework an effective solution for real-world manufacturing FL applications.

\subsection{Limitations and future work}

Our study highlights the critical importance of both the volume of data and the approach to integrating data in FL implementations within industrial settings. It also points out the necessity for additional model optimization to improve its ability to handle unbalanced data distributions, particularly in more complex scenarios like the Contam domain. Future research should focus on adaptive learning strategies or domain-specific adjustments to manage the challenges posed by unbalanced data better, thereby enhancing the model's overall effectiveness.

Additionally, the approach to privacy in our current algorithm is foundational, serving as a solid base for further enhancements. FL clients face various security threats, such as data poisoning, model evasion, model update poisoning, and data inference attacks, which could originate from either participating clients or a malicious server. In the field of FL, techniques like differential privacy, homomorphic encryption, and secure multiparty computation are extensively discussed for protecting client data. These methods for enhancing privacy should be rigorously examined and potentially incorporated into our framework, particularly for applications involving real industrial data. Implementing these security measures would not only enhance data protection but also increase the trustworthiness and practicality of the FL model in industrial settings.

\section{Conclusion}\label{sec:conclusion}

The limited availability of data, combined with the ever-changing and varied configurations in modern manufacturing, has posed significant challenges to the application of machine learning, especially in CM. To address these challenges, this paper presents a novel FTL-TP framework to provide enhanced domain generalization capabilities in FL. Through learning a unified representation from feature space, FTL-TP is able to automatically adapt CM models for clients belong to distince domain groups. Implementations on real-world UMW datasets demonstrate that FTL-TP not only outperforms other learning paradigms but also state-of-the-art FL methods. Compared with baseline FL algorithms, FTL-TP has 5.35\%--8.08\% improvement of accuracy domain generalization tasks. FTL-TP is also shown to achieve excellent performance in challenging scenarios involving unbalanced data distributions and limited client fractions. Furthermore, the FTL-TP method is implemented on an edge-cloud architecture. Results show that FTL-TP both viable and efficient in practice. It is worth noting that the FTL-TP framework is readily extensible to various other industrial applications where challenges of data availability and data privacy are present simultaneously.

\section*{Acknowledgments}
This research has been supported by the National Science Foundation under Grant Nos. 1944345 and  2126246.

\section*{CRediT authorship contribution statement}
\textbf{Ahmadreza Eslaminia}: Conceptualization, Methodology, Software, Formal analysis, Investigation, Validation, Visualization, Writing – original draft, Writing – review \& editing. \textbf{Yuquan Meng}: Conceptualization, Formal analysis, Investigation, Data curation, Visualization, Writing – original draft, Writing – review \& editing. \textbf{Klara Nahrstedt}: Conceptualization, Methodology, Resources, Investigation, Visualization, Writing – original draft, Writing – review \& editing, Supervision, Project administration, Funding acquisition. \textbf{Chenhui Shao}: Conceptualization, Methodology, Resources, Investigation, Visualization, Writing – original draft, Writing – review \& editing, Supervision, Project administration, Funding acquisition.

\bibliography{mybibfile}

\appendix
\section{}
\label{appendix:first}

\renewcommand{\thefigure}{A\arabic{figure}}
% Reset the algorithm counter
\setcounter{algorithm}{0}
% Prefix algorithm numbers with "A"
\renewcommand{\thealgorithm}{A\arabic{algorithm}}

\begin{figure}[htbp]
  \centering
  \begin{minipage}{\linewidth}
    \begin{algorithm}[H]
    \caption{Federated Averaging (FedAvg)}
    \label{alg:fedavg}
    \begin{algorithmic}[1]
      \State \textbf{Server side:} 
      \State Initialize parameters $\theta^{0}$
      \For{each server round $t = 1, 2, \ldots$}
        \State $m \gets \max(C \cdot M, 1)$
        \State $m_t \gets$ (random set of $m$ clients)
        \State Send the current global model $\theta_{t-1}$ to the selected clients
        \For{each client $k \in m_t$ in parallel}
          \State $\theta_k^t \gets$ ClientUpdate($k, \theta^{t-1}$)
          \State Receive updates $\theta_k^t$ from client $k$
        \EndFor
        \State $N \gets \sum_{k=1}^m n_k$ 
        \State $\theta_t \gets \sum_{k=1}^{m}\frac{n_k}{N} \theta_k^t$
      \EndFor
    \end{algorithmic}

    \begin{algorithmic}[1]
    \State \textbf{Client side:} % Mention "Client side" here
    \For{each local epoch $i = 1, 2, \ldots, E$}
      \For{each mini-batch $b$ of size $B$}
        \State $\theta \gets \theta - \eta \nabla \mathcal{F}(\theta, b)$
      \EndFor
    \EndFor
    \State Send the updated local model $\theta$ to the server
    \end{algorithmic}
    \end{algorithm}
  \end{minipage}
\end{figure}

\begin{table}[htbp]
\renewcommand\thetable{A.1}
\centering
\caption{Domain group M client distribution in case of taking Al-Al as target domain (N (New) - W (Worn))}
\label{tab:material_domain}
\begin{tabular}{lcccc}
\hline
Domains/Classes & TC1 (N-N) & TC2 (N-W) & TC3 (W-N) & TC4 (W-N) \\
\hline
Al-Al           & \multicolumn{4}{c}{Target Domains}     \\
Al-Cu           & \multicolumn{4}{c}{200 data points in 3 training clients (about 67 each)} \\
Cu-Cu           & \multicolumn{4}{c}{200 data points in 3 training clients (about 67 each)} \\
Cu-Al           & \multicolumn{4}{c}{200 data points in 3 training clients (about 67 each)} \\
\hline
\end{tabular}
\end{table}

\begin{table}[htbp]
\renewcommand\thetable{A.2}
\centering
\caption{Domain group S client distribution in case of taking clean as target domain}
\label{tab:surface_domain_distribution}
\begin{tabular}{lccc}
\hline
Domain/Classes & DMGD  \hspace{10mm}  & New & Worn \\
\hline
Clean           & \multicolumn{3}{c}{Target domain} \\
Contam    & \multicolumn{3}{c}{90 data points in 3 training clients (30 each)} \\
Polished        & \multicolumn{3}{c}{90 data points in 3 training clients (30 each)} \\
\hline
\end{tabular}
\end{table}

\begin{table}[htbp]
\renewcommand\thetable{A.3}
\centering
\caption{Domain group T client distribution in case of taking DMGD as target domain}
\label{tab:tool_domain_distribution}
\begin{tabular}{lccc}
\hline
Domain/Classes & Clean \hspace{7mm} & Contam & Polished \\
\hline
DMGD  & \multicolumn{3}{c}{Target domain} \\
New   & \multicolumn{3}{c}{90 data points in 3 training clients (30 each)} \\
Worn  & \multicolumn{3}{c}{90 data points in 3 training clients (30 each)} \\
\hline
\end{tabular}
\end{table}

\end{document}